# Cross Language Text Classification via Subspace Co-Regularized Multi-View Learning

**Yuhong Guo**  yuhong@temple.edu
**Min Xiao**  minxiao@temple.edu
Department of Computer and Information Sciences, Temple University, Philadelphia, PA 19122 USA

## Abstract

In many multilingual text classification problems, the documents in different languages often share the same set of categories. To reduce the labeling cost of training a classification model for each individual language, it is important to transfer the label knowledge gained from one language to another language by conducting cross language classification. In this paper we develop a novel subspace co-regularized multi-view learning method for cross language text classification. This method is built on parallel corpora produced by machine translation. It jointly minimizes the training error of each classifier in each language while penalizing the distance between the subspace representations of parallel documents. Our empirical study on a large set of cross language text classification tasks shows the proposed method consistently outperforms a number of inductive methods, domain adaptation methods, and multi-view learning methods.

## 1. Introduction

With the rapid growth of multilingual data in all aspects of human society, it is very common that documents in different languages share the same set of categories. In such multilingual learning scenarios, applying standard monolingual classification methods directly requires costly and time-consuming document annotation in each language. Thus developing effective cross language text classification methods, which transfer the categorization knowledge in a label-rich language, *source language*, to assist classifications in a label-scarce language, *target language*, is becoming increasingly important.

Previous work on cross language text classification mainly focuses on the use of automatic machine translation technology. Most of these methods translate documents from the source language to the target language or vice versa, and then apply standard monolingual classification methods. However, due to the difference in language and culture, there exists a word drift problem. That is, while a word frequently appears in one language, its translated version may rarely appear in the other language. This creates a data distribution discrepancy between the translated training documents from the source language and the original testing documents in the target language, which poses a standard domain adaptation problem. Although many domain adaptation methods can be used in cross language text classification on the top of machine translation, e.g., the work in (Shi et al., 2010; Wei & Pal, 2010; Prettenhofer & Stein, 2010; Wan et al., 2011), they nevertheless suffer from the information loss and translation error introduced in machine translation process without direct access to the original documents. Multi-view learning methods on the other hand treat each language as one independent view of the data and use both the translated documents and the original documents in each language for text classification (Wan, 2009; Amini et al., 2009; Amini & Goutte, 2010).

In this paper, we propose a novel subspace co-regularized multi-view learning method to address cross language text classification based on machine translation. Our assumption is that a document and its translated version describe the same data object in two different views. The underlying discriminative subspace representations of the same data object in the two views thus should be very similar regarding the same classification task. We then simultaneously train two different classifiers, one for each language, by formulating a semi-supervised optimization prob-lem that minimizes the training losses on the labeled data in both views and penalizes the distance between the two projected subspace representations of all data objects. We develop a gradient descent optimization algorithm with curvilinear search to solve the proposed optimization problem for a local optimal solution. Our extensive empirical study on a large number of cross language text classification tasks suggests the proposed approach consistently outperforms a number of comparison inductive methods, domain adaptation methods, and multi-view learning methods.

## 2. Related Work

Previous work on *cross language text classification* mostly relied on machine translation methods, by translating the test data into the language of the training data or vice versa, so that classification algorithms for monolingual texts can be applied (Bel et al., 2003; Shanahan et al., 2004). Although simple and intuitive, these methods suffer from the error and noise introduced in machine translation and the discrepancy of data distribution across languages (Shi et al., 2010). Various methods have been proposed to tackle these issues on translated data to increase cross language text classification accuracy, including an information bottleneck method (Ling et al., 2008), EM-based model translation techniques (Shi et al., 2010; Rigutini & Maggini, 2005), and cross language domain adaptation methods (Wei & Pal, 2010; Prettenhofer & Stein, 2010; Wan et al., 2011).

*Domain adaptation* refers to the problem of adapting a prediction model trained on data from a source domain to a different target domain, where the data distributions in the two domains are different. Effective domain adaptation techniques are essential when labeled data are scarce or barely available in the target domain while there are plenty labeled instances in the source domain. A major challenge of domain adaptation is the data distribution divergence between the source and target domains. Some domain adaptation methods attempt to bridge the distribution gap between the two domains by conducting instance weighting (Bickel et al., 2007) or co-training (Chen et al., 2011). Many others propose to reduce the domain divergence by learning generalizable features from the two domains, including structural correspondence learning (Blitzer et al., 2006), coupled subspace learning (Blitzer et al., 2011), and feature augmentation methods, easyadapt (EA) (Daumé III, 2007) and its co-regularization based semi-supervised extension (EA++) (Daumé III et al., 2010). These methods however are unsuitable for domain adaptation tasks where the feature spaces of the two domains are different. Nevertheless, with machine translation, the cross language text classification problem naturally forms a domain adaptation problem, where the source domain includes the documents translated from the source language and the target domain includes the original documents in the target language. The domain divergence in this case mainly comes from the word drift due to the differences of culture and linguistic expression in different language regions. Many existing domain adaptation methods can be used for cross-lingual text classification. (Wei & Pal, 2010; Prettenhofer & Stein, 2010) use structural correspondence learning for cross language text classification. (Wan et al., 2011) presents a feature and instance bi-weighting adaptation method for cross language text classification. These domain adaptation methods nevertheless can not directly exploit the original documents existing in the source language and thus suffer from the information loss introduced in machine translation process.

Recently, *multi-view learning* methods in combination with machine translation have been applied on multilingual learning scenarios, including cross language text classification tasks. Using machine translation, documents in each language can be translated into parallel documents in the other language to create two independent views of the text objects in different feature spaces. A few multi-view learning methods then have been applied on such multi-view data, including the co-training method (Wan, 2009) which is an instance of the standard co-training algorithm of (Blum & Mitchell, 1998), the multi-view majority voting method (Amini et al., 2009), and the multi-view co-classification method (Amini & Goutte, 2010) which is an instance of the co-regularized multi-view classification (Sindhwani et al., 2002; Sindhwani & Rosenberg, 2008). These multi-view learning methods can exploit the original documents in both languages without translation information loss.

## 3. Cross Language Text Classification

In this work, we combine the domain adaptation intuition of learning generalizable feature representations with the co-regularization principle of multi-view learning, and develop a subspace co-regularized multi-view method for cross language text classification. For simplicity, we consider binary classification tasks. We assume there are documents in two languages, the source language and the target language, for the same classification task. We exploit the data in the label-rich source language to assist training classifiers for the data in the label-scarce target language on top of ma-chine translation. In this section, we first introduce the basic notations, and then present the proposed multi-view learning method.

## 3.1. Notations

Assume there are $n_s$ documents in the source language where $l_s$ of them are labeled and the remaining $u_s$ documents are unlabeled. Similarly, assume there are $n_t$ documents in the target language where $l_t$ ($l_t < l_s$) of them are labeled and the remaining $u_t$ documents are unlabeled. Using machine translation, we can translate each document in the source language into a parallel document in the target language, and vice versa. Combing the original and translated data together in each language, we obtain two parallel matrices, $X_1 \in \mathbb{R}^{n \times d_1}$ in the source view and $X_2 \in \mathbb{R}^{n \times d_2}$ in the target view, where $n = n_s + n_t$. The first $l = l_s + l_t$ rows of $X_1$ and $X_2$ form the labeled submatrices, $X_1^\ell$ and $X_2^\ell$, respectively. Their corresponding labels are given as a column vector $\mathbf{y} \in \{-1, +1\}^l$.

## 3.2. Multi-View Training via Subspace Co-Regularization

We assume there is a low-dimensional subspace representation of the data in each view. The linear predictive function in the $i$th view is derived from the subspace as follows

$$f_i(X_i^\ell) = X_i^\ell \Theta_i \mathbf{w}_i + b_i \quad (1)$$

where $\mathbf{w}_i \in \mathbb{R}^m$ is the linear weight vector, $b_i \in \mathbb{R}$ is the bias parameter, $\Theta_i \in \mathbb{R}^{d_i \times m}$ is the linear transformation matrix that projects the input data into the low-dimensional subspace, and $m$ is the dimensionality of the subspace. The transformation matrix $\Theta_i$ has orthogonal columns such that $\Theta_i^\top \Theta_i = I$ where $I$ is an identity matrix. Since the same classification task is shared between the two views, the underlying predictive subspace representations of the parallel documents in the two views should be very similar. We thus formulate the cross language text classification as a semi-supervised multi-view optimization problem that minimizes the training losses on the labeled data in each view while penalizing the distance between the two view subspace representations of both labeled and unlabeled data. Specifically, we conduct training by minimizing the following regularized loss over the model parameters $\{\Theta_i, \mathbf{w}_i, b_i\}_{i=1}^2$,

$$\sum_{i=1}^2 \mathcal{V}(f_i(X_i^\ell), \mathbf{y}) + \alpha_i \|\mathbf{w}_i\|^2 + \gamma d(X_1 \Theta_1, X_2 \Theta_2),$$

subject to the constraints $\Theta_1^\top \Theta_1 = I$ and $\Theta_2^\top \Theta_2 = I$. Here $\mathcal{V}(\cdot, \cdot)$ is a general loss function, $d(\cdot, \cdot)$ is a distance function that measures the distance between the two projected low-dimensional matrices, $\{\alpha_i\}_{i=1}^2$ and $\gamma$ are tradeoff parameters. By conducting two-view semi-supervised training, we expect the subspace representations can capture both the task specific discriminative information of the labeled data and the underlying intrinsic information of the unlabeled data.

In this work, we consider a least square loss function and a squared Euclidean distance function, i.e.,

$$\mathcal{V}(f_i(X_i^\ell), \mathbf{y}) = \|X_i^\ell \Theta_i \mathbf{w}_i + b_i - \mathbf{y}\|^2 \quad (2)$$
$$d(X_1 \Theta_1, X_2 \Theta_2) = \|X_1 \Theta_1 - X_2 \Theta_2\|_F^2 \quad (3)$$

where $\|\cdot\|_F^2$ denotes the Frobenius norm of matrix. Hence we get the following optimization problem

$$\begin{aligned} \min_{\{\Theta_i, \mathbf{w}_i, b_i\}} \quad & \sum_{i=1}^2 \|X_i^\ell \Theta_i \mathbf{w}_i + b_i - \mathbf{y}\|^2 + \alpha_i \|\mathbf{w}_i\|^2 \\ & + \gamma \|X_1 \Theta_1 - X_2 \Theta_2\|_F^2 \quad (4) \\ \text{s. t.} \quad & \Theta_1^\top \Theta_1 = I, \quad \Theta_2^\top \Theta_2 = I. \end{aligned}$$

Below we show that the optimal $\{\mathbf{w}_i, b_i\}$ can be solved in terms of $\Theta_1$ and $\Theta_2$ from the optimization problem.

**Lemma 1** *The optimal $\{\mathbf{w}_i^*, b_i^*\}_{i=1}^2$ that solve the optimization problem in Eq. (4) is given by*

$$\mathbf{w}_i^* = (\Theta_i^\top X_i^{\ell\top} H X_i^\ell \Theta_i + \alpha_i I)^{-1} \Theta_i^\top X_i^{\ell\top} H \mathbf{y} \quad (5)$$
$$b_i^* = \frac{1}{l} \mathbf{1}^\top (\mathbf{y} - X_i^\ell \Theta_i \mathbf{w}_i^*) \quad (6)$$

*for $i = 1, 2$, where $H = I - \frac{1}{l} \mathbf{1}\mathbf{1}^\top$ and $\mathbf{1}$ denotes a column vector of length $l$ with all 1 entries.*

**Proof:** Taking the derivatives of the objective function in Eq. (4) with respect to $b_1$ and $b_2$ respectively, and setting them to zeros, we obtain

$$b_i = \frac{1}{l} \mathbf{1}^\top (\mathbf{y} - X_i^\ell \Theta_i \mathbf{w}_i)$$

for $i = 1, 2$. Substituting them back into Eq. (4), we have a new objective function as below

$$\sum_{i=1}^2 \left\| H(X_i^\ell \Theta_i \mathbf{w}_i - \mathbf{y}) \right\|^2 + \alpha_i \|\mathbf{w}_i\|^2 + \gamma \left\| X_1 \Theta_1 - X_2 \Theta_2 \right\|_F^2$$

Then taking derivatives of this new objective function with respect to $\mathbf{w}_1$ and $\mathbf{w}_2$, and setting them to zeros, we obtain

$$\mathbf{w}_i = (\Theta_i^\top X_i^{\ell\top} H X_i^\ell \Theta_i + \alpha_i I)^{-1} \Theta_i^\top X_i^{\ell\top} H \mathbf{y}$$

for $i = 1, 2$. $\square$

Following Lemma 1, the objective function in Eq. (4) can be rewritten as below by replacing $\{\mathbf{w}_i, b_i\}$

$$L(\Theta_1, \Theta_2) = \gamma \|X_1\Theta_1 - X_2\Theta_2\|_F^2 + 2\mathbf{y}^\top H\mathbf{y} \quad (7)$$
$$- \sum_{i=1}^{2} \mathbf{z}_i^\top \Theta_i (\Theta_i^\top M_i \Theta_i + \alpha_i I)^{-1} \Theta_i^\top \mathbf{z}_i$$

where $M_i$ and $\mathbf{z}_i$ are defined as

$$M_i = X_i^{\ell\top} H X_i^\ell \quad \text{and} \quad \mathbf{z}_i = X_i^{\ell\top} H\mathbf{y}.$$

Hence the optimization problem in Eq. (4) can be equivalently re-expressed as

$$\min_{\Theta_1, \Theta_2} L(\Theta_1, \Theta_2) \quad \text{s. t.} \quad \Theta_1^\top \Theta_1 = I, \Theta_2^\top \Theta_2 = I. \quad (8)$$

The problem above is a non-convex optimization problem. Nevertheless, the gradient of the objective function with respect to $\{\Theta_1, \Theta_2\}$ can be easily computed, and its part corresponding to each $\Theta_i$ is given as

$$\begin{aligned}\nabla_{\Theta_i} L(\Theta_1, \Theta_2) &= 2\gamma X_i^\top (X_i\Theta_i - X_{\bar{i}}\Theta_{\bar{i}}) \quad (9)\\ &\quad - 2\mathbf{z}_i \mathbf{z}_i^\top \Theta_i (\Theta_i^\top M_i \Theta_i + \alpha_i I)^{-1} \\ &\quad + 2M_i \Theta_i (\Theta_i^\top M_i \Theta_i + \alpha_i I)^{-1} \\ &\quad\quad \Theta_i^\top \mathbf{z}_i \mathbf{z}_i^\top \Theta_i \ (\Theta_i^\top M_i \Theta_i + \alpha_i I)^{-1}\end{aligned}$$

for $\{i=1, \bar{i}=2\}$ or $\{i=2, \bar{i}=1\}$.

### 3.3. Optimization Algorithm

The non-convex optimization problem (8) is generally difficult to optimize due to the orthogonal constraints. In this work, we use a gradient descent optimization procedure with curvilinear search (Wen & Yin, 2010) to solve it for a local optimal solution.

In each iteration of the gradient descent procedure, given the current feasible point $(\Theta_1, \Theta_2)$, the gradients can be computed using (9), such that

$$G_1 = \nabla_{\Theta_1} L(\Theta_1, \Theta_2), \quad G_2 = \nabla_{\Theta_2} L(\Theta_1, \Theta_2). \quad (10)$$

We then compute two skew-symmetric matrices

$$F_1 = G_1 \Theta_1^\top - \Theta_1 G_1^\top, \quad F_2 = G_2 \Theta_2^\top - \Theta_2 G_2^\top. \quad (11)$$

It is easy to see $F_1^\top = -F_1$ and $F_2^\top = -F_2$. The next new point can be searched as a curvilinear function of a step size variable $\tau$, such that

$$Q_1(\tau) = \left(I + \frac{\tau}{2}F_1\right)^{-1}\left(I - \frac{\tau}{2}F_1\right)\Theta_1 \quad (12)$$
$$Q_2(\tau) = \left(I + \frac{\tau}{2}F_2\right)^{-1}\left(I - \frac{\tau}{2}F_2\right)\Theta_2 \quad (13)$$

It is easy to verify that $Q_1(\tau)^\top Q_1(\tau) = I$ and $Q_2(\tau)^\top Q_2(\tau) = I$ for all $\tau \in \mathbb{R}$. Thus we can stay in the feasible region along the curve defined by $\tau$. Moreover, $\frac{d}{d\tau}Q_1(0)$ and $\frac{d}{d\tau}Q_2(0)$ are equal to the projections of $(-G_1)$ and $(-G_2)$ onto the tangent space $\mathcal{Q} = \{(\Theta_1, \Theta_2) : \Theta_1^\top \Theta_1 = I, \Theta_2^\top \Theta_2 = I\}$ at the current point $(\Theta_1, \Theta_2)$. Hence $\{Q_1(\tau), Q_2(\tau)\}_{\tau \geq 0}$ is a descent path in the close neighborhood of the current point. We thus apply a similar strategy as the standard backtracking line search to find a proper step size $\tau$ using curvilinear search, while guaranteeing the iterations to converge to a stationary point. We determine a proper step size $\tau$ as one satisfying the following Armijo-Wolfe conditions

$$\begin{aligned}L(Q_1(\tau), Q_2(\tau)) &\leq L(Q_1(0), Q_2(0)) \quad (14) \\ &\quad + \rho_1 \tau L'_\tau(Q_1(0), Q_2(0)), \\ L'_\tau(Q_1(\tau), Q_2(\tau)) &\geq \rho_2 L'_\tau(Q_1(0), Q_2(0)) \quad (15)\end{aligned}$$

Here $L'_\tau(Q_1(\tau), Q_2(\tau))$ is the derivative of $L$ with respect to $\tau$,

$$L'_\tau(Q_1(\tau), Q_2(\tau)) \quad (16)$$
$$= -\sum_{i=1}^{2} tr\left(R_i(\tau)^\top \left(I + \frac{\tau}{2}F_i\right)^{-1} F_i\left(\frac{\Theta_i + Q_i(\tau)}{2}\right)\right)$$

where $R_i(\tau) = \nabla_{\Theta_i} L(Q_1(\tau), Q_2(\tau))$. Therefore

$$\begin{aligned}L'_\tau(Q_1(0), Q_2(0)) &= -\sum_{i=1}^{2} tr\left(G_i^\top (G_i \Theta_i^\top - \Theta_i G_i^\top)\Theta_i\right) \\ &= -\frac{1}{2}\|F_1\|_F^2 - \frac{1}{2}\|F_2\|_F^2 \quad (17)\end{aligned}$$

The overall algorithm is given in Algorithm 1.

### 3.4. Multi-View Testing

After the semi-supervised multi-view training, we obtain two prediction models defined in Eq. (1) with model parameters $\{\Theta_i, \mathbf{w}_i, b_i\}_{i=1}^{2}$. We then conduct multi-view testing on new documents. Specifically, given a test document, $\mathbf{x} \in \mathbb{R}^{d_2}$, in the target language, we first translate it into the source language to obtain $\widehat{\mathbf{x}} \in \mathbb{R}^{d_1}$. Then we compute the prediction values using the two prediction models

$$f_1(\widehat{\mathbf{x}}) = \widehat{\mathbf{x}}^\top \Theta_1 \mathbf{w}_1 + b_1, \quad (18)$$
$$f_2(\mathbf{x}) = \mathbf{x}^\top \Theta_2 \mathbf{w}_2 + b_2. \quad (19)$$

The prediction confidence of each predictor can be calculated as $|f_1(\widehat{\mathbf{x}})|$ and $|f_2(\mathbf{x})|$ respectively. We finally set the prediction label for $\mathbf{x}$ as the one predicted from the most confident predictor, i.e.,

$$y = \begin{cases} \text{sign}(f_1(\widehat{\mathbf{x}})) & \text{if } |f_1(\widehat{\mathbf{x}})| > |f_2(\mathbf{x})| \\ \text{sign}(f_2(\mathbf{x})) & \text{otherwise} \end{cases} \quad (20)$$

**Algorithm 1** Optimization procedure
  **Input:** $\epsilon \geq 0$ , $0 < \mu < 1$, $0 < \rho_1 < \rho_2 < 1$,
    and initial feasible $\Theta_1, \Theta_2$.
  **Procedure**
  **for** $iter = 1$ to $maxiters$
    1. compute gradients $G_1, G_2$ using Eq. (9),(10).
    2. **if** $\|G_1\|_F^2 + \|G_2\|_F^2 \leq \epsilon$ **then** stop and exit.
    3. compute $F_1$, and $F_2$ according to Eq. (11).
    4. compute $L'_\tau(Q_1(0), Q_2(0))$ via Eq. (17)
    5. set $\tau = 1$
    6. **for** $s = 1$ to $maxsteps$
      -compute $Q_1(\tau), Q_2(\tau)$ via Eq. (12)(13)
      -compute $L'_\tau(Q_1(\tau), Q_2(\tau))$ via Eq. (16)
      -**if** Armijo-Wolfe conditions in Eq. (14) (15)
        are satisfied **then** break-out
      -set $\tau = \mu\tau$
    **end for**
    7. **if** $s > maxsteps$ **then** stop and exit.
    7. update $\Theta_1 = Q_1(\tau)$, $\Theta_2 = Q_2(\tau)$.
  **end for**

## 4. Experiments

In this section, we report our empirical results on a set of cross language text classification tasks.

### 4.1. Experimental Setting

The experiments were conducted on cross language text classification (CLTC) tasks constructed from a comparable multilingual corpus used in (Amini et al., 2009), which contains newswire articles written in 5 languages (English(E), French(F), German(G), Italian(I), Spanish(S)), distributed over 6 classes (*C15, CCAT, E21, ECAT, GCAT, M11*). In this multilingual corpus, each original document was translated into the other 4 languages using a statistical machine translation system. Our *first* set of experiments aim to evaluate CLTC tasks with different languages. We constructed a set of 20 binary cross language classification tasks over all possible source-target pairs of 5 languages, using two large classes, CCAT and ECAT, as shown in Table 1. For example, **E2F** denotes the task that uses *English* as the source language and uses *French* as the target language. For each language, we randomly selected 2000 original documents for each class to form the datasets. Thus in each task, we used 4000 original documents and 4000 translated documents in each language. In each language, we used the top 400 features according to the sums of their TFIDF weights over all documents.

*Next*, we constructed datasets to evaluate CLTC tasks on different classes. We selected 3 languages, French(F), German(G) and Italian(I), that have sufficient number of documents in all 6 classes to use. We then constructed 36 1-vs-all binary classification problems over all 6 classes using 6 source-target pairs of languages, as shown in Table 2. For example, **F2G** denotes the tasks that use *French* as the source language and use *German* as the target language; *C15* denotes the 1-vs-all binary classification task on class *C15*. For each task, we randomly selected 2000 original documents from both the target class and the remaining classes in each language. Thus same as above, we used 4000 original documents and 4000 translated documents for each task in each language.

In the experiments, we compared the proposed Subspace Co-regularized Multi-View learning method (**SCMV**) method with five other methods: (1) **TB**, a baseline method that trains a classifier using only the labeled original documents in the target language; (2) **TSB**, a baseline method that trains a classifier on both the labeled original documents in the target language and the labeled documents translated from the source language; (3) **EA++**, the co-regularization based semi-supervised domain adaptation method developed in (Daumé III et al., 2010), which uses a synthetic source domain formed by translating all documents in the source language into the target language; (4) **MVMV**, the multi-view majority voting method developed in (Amini et al., 2009); and (5) **MVCC**, the semi-supervised version of the multi-view co-classification method (Amini & Goutte, 2010), which penalizes the disagreement of the two view predictions on unlabeled data. Among these methods, only the MVCC uses a logistic regression predictor as base classifier, and all other methods use least squares predictors as base classifiers.

### 4.2. Experiment I

The first set of experiments are conducted on the 20 CLTC tasks constructed above. For each task, we randomly chose 900 labeled and 2100 unlabeled original documents from the source language domain, and chose 100 labeled and 2900 unlabeled original documents from the target language domain for classification model training. Thus in total we had 1000 labeled documents and 5000 unlabeled documents in each language view for training. We used the remaining 1000 original documents in the target language as testing data. Based on this random data partition procedure, we repeated the E2F experiment 3 times to conduct model parameter selection for *MVCC* and the proposed *SCMV*. The trade-off parameter for *MVCC* is selected from $\{1/10, 1/2, 1\}$. For the proposed *SCMV*, we used $\alpha_1 = 0.1, \alpha_2 = 0.1$ and selected $\gamma$ from $\{1/600, 1/60, 1/30, 1/12, 1/6\}$. Finally we selected $1/2$ as the trade-off parameter for MVCC and selected $1/6$ as the $\gamma$ parameter for SCMV.

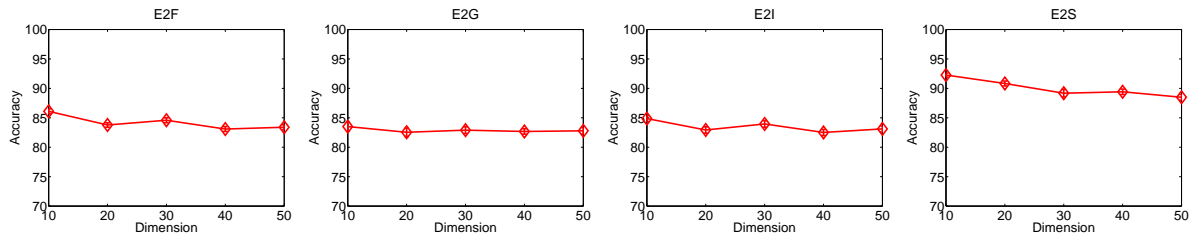

Figure 1. Average classification accuracy results for the SCMV method with different subspace dimensions.

Using the selected parameters and the stated random data partition procedure, we then set the subspace dimension for SCMV as 10, and repeated each experiment 10 times for the six methods in consideration. The average classification results on the test data in term of accuracy are reported in Table 1. We can see that between the two baseline methods, by exploiting the translated labeled documents from the source language, TSB has slight advantages over TB on many tasks. The domain adaptation method, EA++, however, produced similar performance as the baseline TSB. By exploiting both original data and translated data in the two languages, even the simple multi-view method, MVMV, works well on most tasks except the G2I and G2S. The semi-supervised multi-view co-classification method, MVCC, consistently outperforms the four methods mentioned above, although the improvements are not significant on a few tasks including E2G, I2G and I2S. The proposed SCMV on the other hand consistently outperforms the other five methods on all tasks. The improvements over the first four methods are significant across all tasks. Even comparing to MVCC, the improvements are significant over most tasks.

**Subspace dimensions.** Next, we empirically studied the influence of subspace dimension size on the proposed SCMV method. We repeated the experiments above for SCMV with different subspace dimension sizes, m=$\{10,20,30,40,50\}$. The average test accuracy results on tasks E2F, E2G, E2I and E2S are reported in Figure 1. We can see the performance of SCMV is not very sensitive to the subspace dimension size within the considered dimension range.

### 4.3. Experiment II

The second set of experiments are conducted on the 36 CLTC tasks constructed with 1-vs-all classification problems. We used the same random data partition procedure and model parameters stated in experiment I. For the proposed method SCMV, we used 10 as the subspace dimension size. We repeated each experiment 10 times and the average test accuracy results are reported in Table 2. Similar to the first set of experiments, the proposed approach consistently outperforms all the other five methods on all 36 tasks, and the improvements are significant on 22 tasks comparing to the best comparison results.

All these results suggest that identifying two-view consistent subspace representations based on both the original and translated data in two languages can effectively overcome the cross language divergence and achieve good prediction models.

## 5. Conclusion

In this paper, we proposed a novel subspace co-regularized multi-view learning method to address cross language text classification. By training two subspace based prediction models in two language views together while penalizing the distance between the two projected subspace representations of both labeled and unlabeled instances, the underlying discriminative subspace representations can be identified to produce prediction models with better generalization performance. We developed a gradient descent algorithm with curvilinear search to solve the proposed joint optimization problem for a local optimal solution. Our extensive empirical results on a large number of cross language text classification tasks demonstrated the superior performance of the proposed method comparing to a few inductive methods, domain adaptation methods, and multi-view learning methods.

## References

Amini, M. and Goutte, C. A co-classification approach to learning from multilingual corpora. *Machine Learning*, 79:105–121, 2010.

Amini, M., Usunier, N., and Goutte, C. Learning from multiple partially observed views - an application to multilingual text categorization. In *Advances in*

Table 1. Average classification accuracy results over 10 runs for 20 CLTC tasks.

| TASKS | TB | TSB | EA++ | MVMV | MVCC | SCMV |
|---|---|---|---|---|---|---|
| E2F | 78.60±0.80 | 79.24±0.51 | 79.52±0.47 | 81.13±0.46 | 83.20±0.38 | **86.10**±0.42 |
| E2G | 75.65±0.67 | 75.01±0.51 | 75.25±0.46 | 80.37±0.76 | 81.62±0.54 | **83.51**±0.74 |
| E2I | 79.80±0.69 | 76.39±0.98 | 76.48±1.02 | 80.01±0.69 | 83.75±0.64 | **84.87**±0.51 |
| E2S | 84.54±1.52 | 85.24±1.01 | 85.43±1.03 | 86.30±0.69 | 89.98±0.42 | **92.26**±0.34 |
| F2E | 77.04±0.92 | 80.32±0.47 | 80.60±0.48 | 81.15±0.44 | 82.51±0.36 | **83.86**±0.35 |
| F2G | 76.41±0.92 | 76.32±0.62 | 76.68±0.49 | 79.66±0.91 | 81.84±0.76 | **83.16**±0.70 |
| F2I | 78.32±0.82 | 77.02±0.78 | 78.87±0.75 | 79.53±0.63 | 82.98±0.47 | **83.25**±0.43 |
| F2S | 84.77±1.05 | 86.24±0.71 | 86.90±0.69 | 87.53±0.68 | 90.96±0.44 | **92.81**±0.25 |
| G2E | 77.04±0.88 | 78.57±0.37 | 78.42±0.36 | 78.68±0.68 | 80.52±0.50 | **82.52**±0.47 |
| G2F | 75.93±0.70 | 77.08±0.51 | 77.22±0.42 | 77.99±0.61 | 80.57±0.48 | **83.55**±0.36 |
| G2I | 79.88±0.77 | 78.54±1.05 | 78.61±0.99 | 78.07±0.78 | 81.85±0.54 | **84.20**±0.51 |
| G2S | 85.82±0.91 | 86.22±0.55 | 86.61±0.57 | 84.73±0.62 | 89.24±0.37 | **90.67**±0.61 |
| I2E | 76.98±0.74 | 76.76±0.42 | 77.80±0.40 | 78.86±0.61 | 80.45±0.47 | **81.34**±0.48 |
| I2F | 76.88±0.94 | 78.10±0.35 | 78.61±0.47 | 78.11±0.65 | 80.58±0.60 | **81.73**±0.42 |
| I2G | 76.79±0.57 | 76.56±0.55 | 77.66±0.48 | 79.69±0.61 | 80.50±0.53 | **84.76**±0.35 |
| I2S | 85.36±1.42 | 87.68±0.50 | 88.63±0.51 | 89.42±0.56 | 90.66±0.33 | **94.15**±0.44 |
| S2E | 74.35±0.94 | 74.73±0.63 | 74.83±0.69 | 77.89±0.54 | 79.45±0.58 | **80.50**±0.44 |
| S2F | 75.89±1.10 | 77.48±0.58 | 77.62±0.57 | 77.93±0.62 | 82.82±0.22 | **84.86**±0.33 |
| S2G | 75.88±0.44 | 74.28±0.40 | 74.31±0.34 | 77.91±0.56 | 80.90±0.44 | **81.12**±0.53 |
| S2I | 79.36±0.84 | 79.72±0.69 | 80.54±0.75 | 82.46±0.65 | 87.18±0.46 | **88.59**±0.47 |

*Neural Information Process. Systems (NIPS)*, 2009.

Bel, N., Koster, C., and Villegas, M. Cross-lingual text categorization. In *Proc. of the European Conference on Digital Libraries (ECDL)*, 2003.

Bickel, S., Brückner, M., and Scheffer, T. Discriminative learning for differing training and test distributions. In *Proc. of the International Conference on Machine Learning (ICML)*, 2007.

Blitzer, J., McDonald, R., and Pereira, F. Domain adaptation with structural correspondence learning. In *Proc. of the Conference on Empirical Methods in Natural Language Processing (EMNLP)*, 2006.

Blitzer, J., Foster, D., and Kakade, S. Domain adaptation with coupled subspaces. In *Proc. of the International Conference on Artificial Intelligence and Statistics (AISTATS)*, 2011.

Blum, A. and Mitchell, T. Combing labeled and unlabeled dta with co-training. In *Proc. of the Annual Conference on Learning Theory (COLT)*, 1998.

Chen, M., Weinberger, K., and Blitzer, J. Co-training for domain adaptation. In *Advances in Neural Information Processing Systems (NIPS)*, 2011.

Daumé III, H. Frustratingly easy domain adaptation. In *Proc. of the Annual Meeting of the Association for Computational Linguistics (ACL)*, 2007.

Daumé III, H., Kumar, A., and Saha, A. Co-regularization based semi-supervised domain adaptation. In *Advances in Neural Information Processing Systems (NIPS)*, 2010.

Ling, X., Xue, G., Dai, W., Jiang, Y., Yang, Q., and Yu, Y. Can chinese web pages be classified with english data source. In *Proc. of the international conference on World Wide Web*, 2008.

Prettenhofer, P. and Stein, B. Cross-language text classification using structural correspondence learning. In *Proc. of the Annual Meeting of the Association for Computational Linguistics (ACL)*, 2010.

Rigutini, L. and Maggini, M. An EM based training algorithm for cross-language text categorization. In *Proc. of the Web Intelligence Conference*, 2005.

Shanahan, J. G., Grefenstette, G., Qu, Y., and Evans, D. A. Mining multilingual opinions through classification and translation. In *Proc. of AAAI'04 Spring Symp. on Explor. Attitude and Affect in Text*, 2004.

Shi, L., Mihalcea, R., and Tian, M. Cross language text classification by model translation and semi-supervised learning. In *Proc. of the Conference on Empirical Methods in Natural Language Processing (EMNLP)*, 2010.

Sindhwani, V. and Rosenberg, D. An RKHS for multi-view learning and manifold co-regularization. In *Proc. of the International Conference on Machine Learning (ICML)*, 2008.

Table 2. Classification accuracy results over 10 runs for 36 CLTC tasks.

| TASKS | CLASS | TB | TSB | EA++ | MVMV | MVCC | SCMV |
|---|---|---|---|---|---|---|---|
| F2G | C15 | 78.58±0.51 | 79.99±0.83 | 80.26±0.96 | 81.00±0.60 | 81.39±0.70 | **83.08**±0.63 |
|  | CCAT | 67.28±0.90 | 67.99±0.46 | 68.57±0.42 | 70.40±0.79 | 71.58±0.90 | **72.80**±0.68 |
|  | E21 | 77.49±0.91 | 76.93±0.82 | 77.60±0.78 | 78.84±0.66 | 79.05±0.63 | **80.59**±0.42 |
|  | ECAT | 64.96±0.81 | 67.29±0.40 | 68.17±0.42 | 70.99±0.69 | 71.65±0.64 | **72.37**±0.74 |
|  | GCAT | 73.09±0.69 | 74.36±0.77 | 74.82±0.71 | 76.65±0.63 | 77.60±0.85 | **80.29**±0.59 |
|  | M11 | 84.13±0.53 | 85.41±0.82 | 85.97±0.75 | 87.77±0.58 | 88.27±0.68 | **90.64**±0.34 |
| F2I | C15 | 75.89±1.12 | 78.92±0.51 | 79.19±0.81 | 80.22±0.42 | 81.05±0.89 | **82.23**±0.50 |
|  | CCAT | 68.29±0.84 | 68.72±0.85 | 69.46±0.81 | 70.55±0.80 | 70.94±0.95 | **72.17**±0.50 |
|  | E21 | 75.16±1.06 | 75.94±0.94 | 76.19±1.02 | 76.31±0.87 | 76.56±0.90 | **77.68**±0.79 |
|  | ECAT | 71.51±1.12 | 72.50±0.71 | 72.74±0.73 | 74.57±0.54 | 74.82±0.64 | **78.20**±0.33 |
|  | GCAT | 71.98±1.51 | 71.84±0.54 | 72.14±0.51 | 73.21±0.62 | 73.75±0.68 | **75.77**±0.65 |
|  | M11 | 88.00±0.61 | 88.28±0.50 | 88.62±0.42 | 88.78±0.28 | 89.35±0.71 | **90.34**±0.28 |
| G2F | C15 | 82.93±0.65 | 83.03±0.65 | 83.10±0.62 | 83.85±0.69 | 84.63±0.94 | **86.03**±1.04 |
|  | CCAT | 71.36±0.96 | 72.59±0.61 | 73.25±0.56 | 74.80±0.63 | 75.57±0.56 | **76.64**±0.45 |
|  | E21 | 80.67±0.57 | 82.74±0.53 | 83.50±0.50 | 84.34±0.57 | 85.17±0.60 | **86.31**±0.55 |
|  | ECAT | 70.45±0.71 | 72.22±0.54 | 72.91±0.48 | 75.75±0.78 | 76.51±0.72 | **78.18**±0.55 |
|  | GCAT | 78.90±0.79 | 81.26±0.60 | 81.25±0.56 | 83.05±0.54 | 83.48±0.50 | **86.39**±0.37 |
|  | M11 | 90.06±0.68 | 90.52±0.84 | 91.31±0.88 | 91.23±0.59 | 92.21±0.55 | **96.29**±0.31 |
| G2I | C15 | 75.99±1.08 | 76.74±0.92 | 77.41±0.81 | 77.76±0.96 | 78.32±0.79 | **79.87**±0.59 |
|  | CCAT | 66.45±1.17 | 67.01±1.04 | 67.60±0.96 | 68.21±0.81 | 68.80±0.80 | **69.96**±0.73 |
|  | E21 | 75.27±1.07 | 75.77±0.70 | 76.06±0.76 | 76.15±0.94 | 76.57±1.17 | **78.60**±0.59 |
|  | ECAT | 72.17±0.58 | 72.43±0.53 | 72.51±0.52 | 73.23±0.94 | 73.62±0.83 | **76.27**±0.42 |
|  | GCAT | 69.35±1.16 | 71.13±0.62 | 71.52±0.54 | 72.06±0.81 | 72.30±0.73 | **72.90**±0.64 |
|  | M11 | 84.19±1.81 | 85.55±0.95 | 85.77±0.88 | 86.56±0.90 | 86.77±0.73 | **89.91**±0.74 |
| I2F | C15 | 83.69±0.81 | 83.44±0.60 | 84.15±0.59 | 86.97±0.57 | 86.91±0.81 | **89.39**±0.47 |
|  | CCAT | 70.99±1.19 | 72.41±0.42 | 73.05±0.50 | 76.83±0.46 | 77.06±0.64 | **78.02**±0.47 |
|  | E21 | 79.65±0.84 | 79.89±0.45 | 80.59±0.44 | 82.19±0.49 | 82.25±0.87 | **83.93**±0.49 |
|  | ECAT | 70.70±0.89 | 72.58±0.76 | 72.84±0.80 | 73.94±0.56 | 74.72±0.68 | **76.31**±0.62 |
|  | GCAT | 77.77±0.65 | 78.02±0.63 | 78.02±0.61 | 81.64±0.43 | 82.12±0.53 | **85.19**±0.37 |
|  | M11 | 92.07±0.73 | 90.23±0.61 | 88.46±0.83 | 93.79±0.79 | 94.15±0.41 | **97.07**±0.25 |
| I2G | C15 | 78.26±0.79 | 78.68±1.12 | 79.49±1.12 | 80.12±0.78 | 80.99±0.91 | **85.95**±0.56 |
|  | CCAT | 66.45±1.05 | 66.97±0.60 | 67.49±0.63 | 70.90±0.84 | 71.61±0.54 | **73.31**±0.69 |
|  | E21 | 75.69±0.92 | 77.11±0.87 | 78.35±0.91 | 77.70±0.50 | 79.81±0.73 | **82.12**±0.37 |
|  | ECAT | 67.34±1.08 | 68.25±0.92 | 69.98±0.96 | 72.57±0.83 | 73.55±0.85 | **74.99**±0.72 |
|  | GCAT | 73.52±0.71 | 74.15±0.78 | 74.27±0.74 | 78.57±0.98 | 80.65±0.62 | **81.04**±0.60 |
|  | M11 | 80.97±0.53 | 80.54±1.40 | 82.59±1.13 | 86.92±1.18 | 89.40±0.66 | **93.57**±0.69 |

Sindhwani, V., Niyogi, P., and Belkin, M. A coregularization approach to semi-supervised learning with multiple views. In *Proc. of the Workshop on Learning with Multiple Views, ICML*, 2002.

Wan, C., Pan, R., and Li, J. Bi-weighting domain adaptation for cross-language text classification. In *Proc. of the International Joint Conference on Artificial Intelligence (IJCAI)*, 2011.

Wan, X. Co-training for cross-lingual sentiment classification. In *Proc. of the Annual Meeting of the Association for Comput. Linguistics (ACL)*, 2009.

Wei, B. and Pal, C. Cross lingual adaptation: an experiment on sentiment classifications. In *Proc. of the Annual Meeting of the Association for Computational Linguistics (ACL)*, 2010.

Wen, Z. and Yin, W. A feasible method for optimization with orthogonality constraints. Technical report, Rice University, 2010.